\documentclass[11pt]{article}

\usepackage[preprint]{acl}

\usepackage{times}
\usepackage{latexsym}

\usepackage[T1]{fontenc}
\usepackage[utf8]{inputenc}

\usepackage{microtype}

\usepackage{inconsolata}

\usepackage{graphicx}

\usepackage{xspace}
\usepackage{setspace}
\usepackage{xcolor}
\definecolor{myblue}{HTML}{0000A5}
\usepackage{wrapfig,lipsum,booktabs}
\usepackage{wrapfig}
\usepackage{subcaption}
\usepackage{caption} 
\usepackage{textcomp}
\usepackage[most]{tcolorbox}
\usepackage{amsmath}
\usepackage{tabularx}
\usepackage{makecell}
\usepackage[super]{nth}
\usepackage{bbding}
\usepackage{diagbox}
\usepackage{enumitem}
\usepackage{amsfonts}
\usepackage{multirow}

\makeatletter
\renewcommand{\eqref}[1]{Eq.~(\ref{#1})}
\makeatother

\newcommand{\ours}{{\normalsize \texttt{SynthAgent}}\xspace}

\title{SynthAgent: Adapting Web Agents with Synthetic Supervision}

\def\myand{\end{tabular}\hss\egroup \hfil\hfil\egroup
           \hbox to \linewidth\bgroup\large \hfil\hfil
             \hbox to 0pt\bgroup\hss \begin{tabular}[t]{c}\bf}

\author{
  Zhaoyang Wang$^{1,3}$\thanks{Work done during internship at Microsoft.}~~Yiming Liang$^{2}$~~Xuchao Zhang$^{3}$~~Qianhui Wu$^{3}$~~Siwei Han$^{1}$\myand 
  Anson Bastos$^{3}$~~Rujia Wang$^{3}$~~Chetan Bansal$^{3}$~~Baolin Peng$^{3}$\myand
  Jianfeng Gao$^{3}$~~Saravan Rajmohan$^{3}$~~Huaxiu Yao$^{1}$\\[0.6em]
  $^{1}$UNC-Chapel Hill~~~$^{2}$Purdue University~~~$^{3}$Microsoft\\
  \texttt{\{zhaoyang,huaxiu\}@cs.unc.edu,\{xuchaozhang,qianhuiwu\}@microsoft.com}
}

\begin{document}

\maketitle

\begin{abstract}
  Web agents struggle to adapt to new websites due to the scarcity of environment specific tasks and demonstrations. Recent works have explored synthetic data generation to address this challenge, however, they suffer from data quality issues where synthesized tasks contain hallucinations that cannot be executed, and collected trajectories are noisy with redundant or misaligned actions. In this paper, we propose \ours, a fully synthetic supervision framework that aims at improving synthetic data quality via dual refinement of both tasks and trajectories. Our approach begins by synthesizing diverse tasks through categorized exploration of web elements, ensuring efficient coverage of the target environment. During trajectory collection, tasks are refined only when conflicts with observations are detected, which mitigates hallucinations while preserving task consistency. After collection, we conduct trajectory refinement with global context to mitigate potential noise or misalignments. Finally, we fine-tune open-source web agents on the refined synthetic data to adapt them to the target environment. Experimental results demonstrate that \ours outperforms existing synthetic data methods, validating the importance of high-quality synthetic supervision.  The code is publicly available at \url{https://github.com/aiming-lab/SynthAgent}.
\end{abstract}

\section{Introduction}
\begin{figure}[t]
  \centering
  \includegraphics[width=\columnwidth]{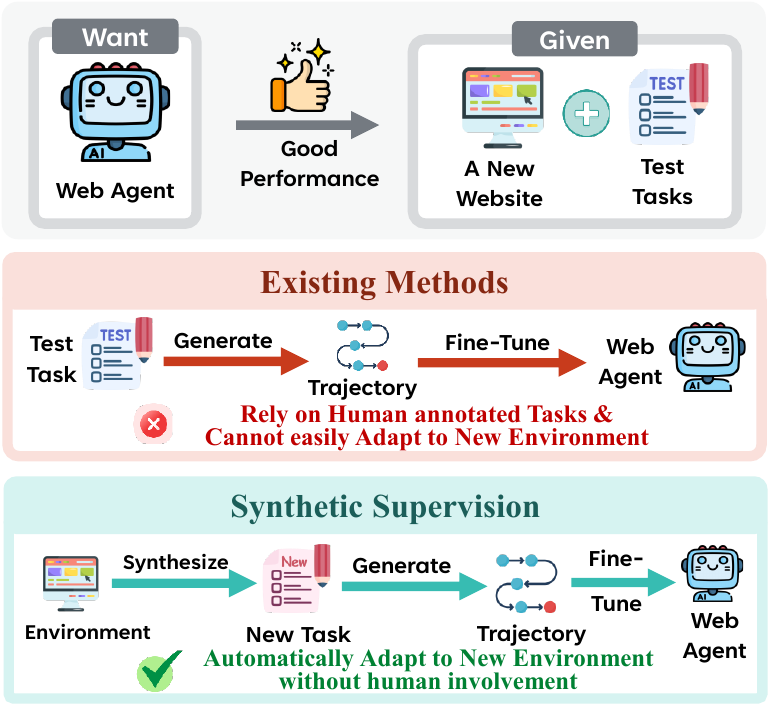}
  \caption{The web agent is expected to adapt and operate on a new environment. Some existing works choose to train on tasks from test set due to the scarcity of environment specific data. In contrast, this paper offers an approach for fully synthetic supervision to adapt agents to such new environment without human involvement.}
  \label{fig:first}
\end{figure}

Large language models (LLMs) with multimodal capabilities have enabled a new wave of web agents capable of autonomously completing complex tasks on the internet~\citep{hong2024metagpt,he2024openwebvoyager,agashe2025agent,li2025websailor,yang2025agentoccam}. These agents take user instructions and then interact with websites to accomplish tasks, showing promising results on standardized benchmarks~\citep{zhou2024webarenarealisticwebenvironment,yao2022webshop,wei2025browsecomp,deng2023mind2webgeneralistagentweb}. However, a persistent challenge is that web agents struggle to adapt to new websites not seen during training~\citep{zhou2024webarenarealisticwebenvironment,pahuja2025explorerscalingexplorationdrivenweb,he2024openwebvoyager}, because new environments often lack sufficient task demonstrations.
And issuing environment specific tasks and collecting trajectories by human on every new website can be expensive. Existing training datasets are either limited to a few domains or lack diversity~\citep{xu2024agenttrek,deng2023mind2webgeneralistagentweb,xie2024osworld,chen2024gui}, thus when an agent is deployed on a new website, it could frequently encounter unfamiliar states or tasks for which it has no experience. This raises the challenge of how to effectively adapt web agents to new environments without human involvement.

A straightforward way to improve an agent's performance on a new website environment is to collect more environment specific training data.
However, traditional agentic data collection relies on human experts or manually scripted tasks~\citep{zhou2024webarenarealisticwebenvironment,lu2024weblinx,deng2023mind2webgeneralistagentweb,mitra2024agentinstruct}. Such approaches are labor-intensive and time consuming, which cannot be easily scaled. 
This may lead to significant gaps between training data and the real-world environments where agents are deployed.
Without abundant experience in the new and unfamiliar environment, it is challenging for agents to adapt and perform effectively.

To address the data scarcity issue, as shown in Figure~\ref{fig:first}, synthetic data~\citep{xu2024agenttrek,su2025learn,sun2025osgenesisautomatingguiagent,pahuja2025explorerscalingexplorationdrivenweb} has emerged as a promising solution, which uses LLMs to generate and collect data for training.
However, existing web agent synthesis pipelines typically optimize only one side of the task-trajectory pair.
Methods like OS-Genesis~\citep{sun2025osgenesisautomatingguiagent} synthesize tasks from single-step observations, which grounds task proposals in real interface changes but provides too little context; as a result, the generated goals often reference non-existent elements or impossible states.
Methods like Explorer~\citep{pahuja2025explorerscalingexplorationdrivenweb} instead keep grounding tasks during execution by continuously refining an underspecified homepage goal, but this can change the task intent after many actions have already been collected, producing trajectories where early segments are misaligned with the final task.
We therefore identify a central design tension in synthetic supervision for web agents: task synthesis requires environment grounding to avoid hallucinations, yet grounding the task aggressively during execution introduces trajectory noise.

In this paper, we propose \ours, a fully synthetic supervision framework that resolves this tension through dual refinement: tasks are refined during execution to mitigate task hallucinations, and trajectories are subsequently refined to mitigate potential noise introduced by task edits.
Specifically, our method consists of four stages: 
(1) First, we synthesize diverse tasks through the proposed categorized exploration that systematically covers functional groups of web elements, improving both task diversity and exploration coverage.
(2) Then, during trajectory collection, we apply task refinement triggered by explicit conflict detection, correcting hallucinations while minimizing unnecessary task changes.
(3) After collection, we conduct trajectory refinement with global context, mitigating noise and misalignments introduced by task edits or agent wandering.
(4) Finally, we fine-tune open-source web agents on the refined synthetic data.
The key insight is that task and trajectory refinement are synergistic: task refinement enables feasible execution but introduces noise, which trajectory refinement subsequently mitigates.

In summary, our contributions are three-fold:
\begin{enumerate}[label=(\arabic*), itemsep=0pt]
\item We propose \ours, a fully synthetic supervision framework for effectively adapting web agents to new environments without test task leakage or human involvement.
\item We identify a critical tension in existing methods: environment grounding during task execution inevitably introduces trajectory noise. We resolve this through a dual refinement strategy. We also propose categorized exploration to enhance task diversity and coverage.
\item Extensive experiments and analyses show that \ours improves synthetic data quality and downstream agent adaptation across benchmarks, highlighting the importance of jointly improving task diversity, task feasibility, and trajectory alignment.
\end{enumerate}

% Moreover, the continuous refinement of tasks along the trajectory can result in a mismatch between the final task description and the originally collected trajectory. Another major issue is that Explorer only focuses on pages that do not require authentication, thus synthesized tasks are often shallow and the exploration coverage remains incomplete.

\section{Related Work}

\subsection{Web Agent}
Recent advances in LLMs have driven interest in developing agents that combine reasoning and interaction~\citep{wei2022chain,yao2023tree,wang2022self,lightman2023let,guo2025deepseek}. 
ReAct~\citep{yao2023reactsynergizingreasoningacting} introduces the interleaving of reasoning and actions after observations, while following works~\citep{gao2023palprogramaidedlanguagemodels,hong2024metagpt,wu2024autogen,nakano2021webgpt,yang2025agentoccam} explore tools integration, planning, and observation-action alignment to enhance agent capabilities. 
However, these typically rely on human annotation, which is costly to scale and adapt to new environments. 
Another series of works~\citep{lu2025ui,qiu2025alita,wei2025webagent} use reinforcement learning to train web agents, but they often require supervision from annotation and are impractical for complex and realistic websites due to the cost of online training. 
Using synthetic data to train web agents has gained traction as a scalable alternative, especially for new environments~\citep{wang2024survey,liu2025advances}.

\subsection{Data Synthesis}
Data synthesis has emerged as a powerful paradigm for addressing data scarcity across various fields, allowing models to learn from automatically generated examples rather than costly human annotations. 
Early works such as Self-Instruct~\citep{wang2023selfinstructaligninglanguagemodels} and Alpaca~\citep{alpaca} leverage advanced but closed-source LLMs to bootstrap instruction tuning data to train smaller open-source models. 
Meanwhile, other studies~\citep{zelikman2022star,wang2023democratizing,ge2024scaling,zhao2025absolute} explore synthesizing large-scale training data to enhance the reasoning performance of smaller LLMs.
These successes demonstrate the promising potential of synthetic data for adapting models to new tasks and domains.
However, in agent scenarios, many works focus solely on synthesizing trajectories while directly using tasks from the test set to train the model~\citep{chen2024agent,zhou2024webarenarealisticwebenvironment,zhang2025symbiotic}, raising serious concerns about test set leakage.
Fortunately, recent works~\citep{he2024openwebvoyager,su2025learn,sun2025osgenesisautomatingguiagent,pahuja2025explorerscalingexplorationdrivenweb} have explored data synthesis for web agents from scratch, including both task and trajectory synthesis.

\subsection{Synthetic data for Web Agents}
Synthetic data has been increasingly used to train web agents~\citep{wang2023selfinstructaligninglanguagemodels,alpaca,xu2023wizardlm,wang2023democratizing}. 
\citet{he2024openwebvoyager} introduce self-instruct for agentic task generation, where it mainly operates on the surface of websites, thus the synthesized tasks are often simple and repetitive. 
AgentTrek~\citep{xu2024agenttrek} collects training data by scraping web tutorials from offline corpus, which can be outdated and not feasible for new environments. 
Synatra~\citep{ou2024synatra} generates offline HTML trajectories by injecting web knowledge from external sources (e.g., wikiHow) without interacting with real websites, which limits the grounding of synthesized data. 
NNetNav~\citep{murty2025nnetnavunsupervisedlearningbrowser} employs an LLM-based trajectory labeler to prune exploration based on instruction hierarchies, but does not address data quality issue.
WebSynthesis~\citep{gao2025websynthesisworldmodelguidedmctsefficient} employs a learned world model of web interfaces to simulate interactions, but its LLM-based environment can introduce additional hallucinations. 
OpenCUA~\citep{wang2025opencua} shows that richer supervision signals annotated by human can substantially improve agent training. 
In contrast, \ours focuses on fully synthetic task and trajectory supervision without human annotation,  integrating additional signals into our synthesis pipeline is a complementary future direction.

% Synthetic data has been increasingly used to train web agents~\citep{wang2023selfinstructaligninglanguagemodels,alpaca,xu2023wizardlm,wang2023democratizing}. 
% \citet{he2024openwebvoyager} apply self-instruct but produce simple and repetitive tasks. 
% AgentTrek~\citep{xu2024agenttrek} scrapes offline web tutorials that can be outdated. 
% Synatra~\citep{ou2024synatra} and NNetNav~\citep{murty2025nnetnavunsupervisedlearningbrowser} generate trajectories from external sources or pruned exploration without addressing data quality.
% WebSynthesis~\citep{gao2025websynthesisworldmodelguidedmctsefficient} simulates interactions via learned world models but may introduce hallucinations. 

Two recent works are most related to ours. 
Explorer~\citep{pahuja2025explorerscalingexplorationdrivenweb} refines underspecified initial tasks during execution, but produces noisy trajectories due to wandering and endless refinement. 
OS-Genesis~\citep{sun2025osgenesisautomatingguiagent} pioneered reverse task synthesis through random GUI exploration, but suffers from hallucinations in single-step-based task proposals and inefficient coverage. In contrast, \ours introduces structural improvements: (1) categorized exploration for systematic coverage and improved task diversity, (2) task refinement that detects and corrects hallucinations based on actual observations, and (3) trajectory refinement with global context that removes noise from task edits or agent wandering.

\begin{figure*}[t]
  \centering
  \includegraphics[width=1.0\textwidth]{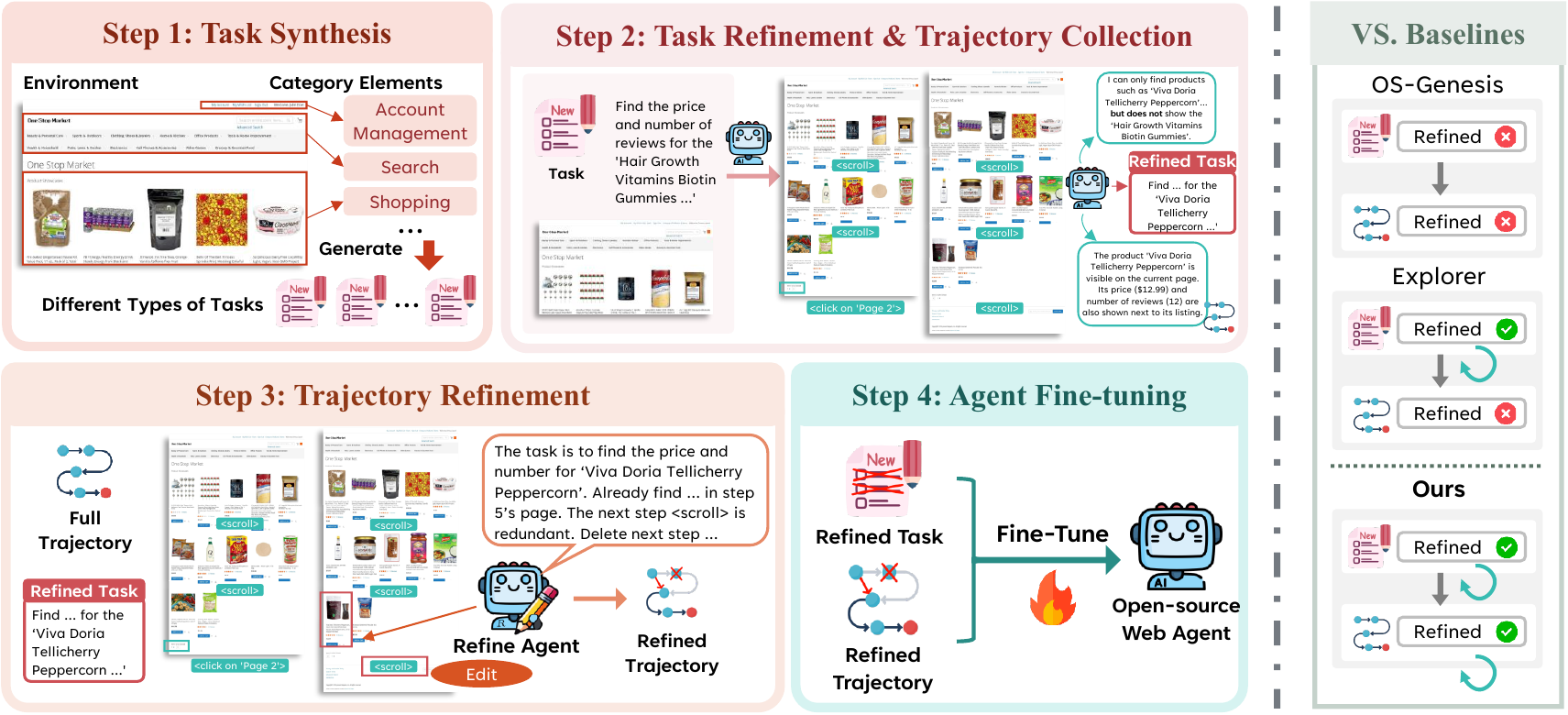}
  \caption{Overview of \ours compared to baseline methods OS-Genesis~\citep{sun2025osgenesisautomatingguiagent} and Explorer~\citep{pahuja2025explorerscalingexplorationdrivenweb}. \ours consists of four steps: (1) Task Synthesis, generating diverse tasks via categorized exploration; (2) Task Refinement, refining tasks during trajectory collection to avoid task  hallucinations; (3) Trajectory Refinement, globally refining collected trajectories to remove noisy actions; and (4) Agent Fine-tuning, adapting web agents using fully synthetic data. The right panel emphasizes the advantage of our dual refinement strategy over baseline methods, leading to higher-quality synthetic web agent data.}
  \label{fig:overview}
\end{figure*}

\section{Method}
As shown in Figure~\ref{fig:overview}, the proposed \ours framework consists of four main steps: (1) Task Synthesis with Categorized Exploration, (2) Task Refinement during Trajectory Collection, (3) Trajectory Refinement, and (4) Agent Fine-tuning. In the following, we describe each step in detail.

\paragraph{Problem Setup.}
We view a website as a partially observable environment $\mathcal{E}$.
The task $\tau$ specifies a goal that requires a web agent to interact with the environment to accomplish.
At step $t$, the agent receives a multimodal observation $o_t$ (i.e., textual accessibility tree~\citep{zhou2024webarenarealisticwebenvironment} and visual screenshot of the webpage) and outputs an action $a_t \in \mathcal{A}$ (e.g., \textsc{Click}, \textsc{Type}, \textsc{Scroll} and etc.).
Let $h_t=(o_1,a_1,\ldots,o_t)$ denote the trajectory, i.e., the sequence of observations and actions taken by the agent to complete the task.
In this paper, our goal is to synthesize a dataset $\mathcal{D}=\{(\tau^i, h^i)\}_{i=1}^N$ of tasks $\tau^{i}$ and corresponding trajectories $h^{i}$ on a previously unseen environment, and then adapt an open-source web agent to this environment by supervised fine-tuning.

\subsection{Task Synthesis}
A key challenge in task synthesis is achieving both diversity (covering different functionalities) and feasibility (avoiding hallucinated goals). Prior work explores environments via random interactions~\citep{sun2025osgenesisautomatingguiagent}, which often revisits redundant elements and leaves important functional regions underexplored. We address this through categorized exploration, which turns raw interface exploration into a function-aware coverage problem: instead of sampling individual elements uniformly, we first group interactive elements by their semantic roles and then explore across these groups.
Specifically, at each visited page $o_t$, we classify interactive elements (buttons, inputs, links, etc.) with their names and roles into functional categories (e.g., Account Management, Search \& Filters, Shopping Content) by prompting an LLM with the page structure. From each category, we uniformly sample up to $2$ unvisited elements to interact with, collecting interaction triplets $(o_t,a_t,o_{t+1})$ where $a_t$ is the corresponding action for the sampled element and $o_{t+1}$ is the resulting page. The per-category sampling budget prevents a single dense page region from dominating exploration, while the unvisited-element constraint encourages broader coverage. We also maintain a URL pool to track discovered pages for deeper exploration.

For each triplet, we prompt the LLM to propose a high-level task $\tau$ that (1) is achievable from $o_t$ through multi-step interactions and (2) is likely completable in $o_{t+1}$ or subsequent pages. Rather than describing the observed action in isolation, the LLM imagines a broader objective where $a_t$ is one grounded step toward the goal. This design links task generation to concrete interface transitions while still allowing multi-step task diversity. Statistics show our exploration yields an average of $6$ functional categories per page with stable granularity across websites, as shown in Table~\ref{tab:cat_stats}.

\subsection{Task Refinement}
\paragraph{Why task refinement is necessary.}
After task proposal, OS-Genesis deploys the agent to interact with the environment to collect trajectories for each synthesized task. Note that its proposed task is based solely on a single interaction, thus it often contains hallucinations (e.g., assuming an option or account state that does not exist), causing the agent to fail to complete the task.

For our method, even with categorized exploration, synthesized tasks may contain hallucinations due to limited observations. A natural solution is to refine tasks during execution when conflicts arise. However, how refinement is triggered matters critically: 
Explorer~\citep{pahuja2025explorerscalingexplorationdrivenweb} combines task proposal and trajectory collection into one stage. It first proposes a coarse task from the homepage and then refines the task continuously at every step. This design implicitly acknowledges that the initial task is underspecified, meaning the following refinement is more about filling in details rather than correcting hallucinations. Empirically, we observe that Explorer frequently changes the task intent during execution ($8.6$ times vs. $2.0$ times of \ours), which often leads to overly long trajectories that fail to complete within the budget ($68.3\%$ samples vs. $6.3\%$ samples of \ours).
In contrast, we design conflict-triggered refinement that activates only when explicit mismatches are detected, resulting in only $2.0$ refinements on average and $6.3\%$ budget exceedance.
The key difference is that our initial tasks are already well-specified, so the refinement should focus on correcting hallucinations rather than continuously changing intention.

\paragraph{Refinement during Trajectory Collection.}
Let $\tau_t$ denote the task specification at time $t$ and $h_t=(o_1,a_1,\ldots,o_t)$ the partial trace. The task is likely to contain hallucinations if it conflicts with the observation on the realistic environment. Thus, we trigger refinement when we detect a conflict with a lightweight predicate $\mathcal{C}(h_t,\tau_t){=}\text{true}$, where
\begin{equation}
  \small
\begin{aligned}
\mathcal{C} := \;&
\underbrace{\neg \textsf{ExistsUI}(h_t,\tau_t)}_{\text{invalid goal}} \\
&\vee\; \underbrace{\textsf{MissingArgs}(h_t,\tau_t)}_{\text{missing details}}
&\vee\; \underbrace{\textsf{Stall}(h_t)}_{\text{stalled}}
\, .
\end{aligned}
\label{eq:stall}
\end{equation}
The first term ``ExistsUI'' fires when element implied by $\tau_t$ are absent or contradict observations (e.g., referenced item does not exist). The second term ``MissingArgs'' detects when $\tau_t$ is underspecified and lacks essential parameters (e.g., username when login) that cannot be inferred from $h_t$. The third term ``Stall'' detects lack of progress: three consecutive no-op transitions, or the same navigation/error loop encountered twice.

Upon a trigger, we call the LLM to follow four evidence-driven principles to refine the task: (1) concretize missing details, (2) align with actual observations, (3) downscope or simplify when blocked, and (4) preserve a similar task category. The replacement task $\tau_{t+1}$ is expected to resolve the conflict while remaining as close as possible to the original intent. The following actions are then taken to continue trajectory collection in order to complete the new task $\tau_{t+1}$. After execution, the final specification is the task $\tau^\star$ along with the collected trajectory $h_T$. Note that the refinement process can introduce discontinuities and noise into the trajectory, which we address in the next step.

\subsection{Trajectory Refinement}
\paragraph{Why Trajectory refinement is necessary.}
Task refinement during execution enables feasible task completion but introduces a side effect: trajectory segments collected under earlier task variants become misaligned with the final task $\tau^\star$. Additionally, even without task edits, agents may wander through dead-ends or repeat no-op actions. This motivates trajectory refinement as a post-hoc step that leverages the global view of the complete trajectory and the final task to remove noise. Critically, task and trajectory refinement are synergistic: the former ensures task feasibility while introducing noise, which the latter subsequently mitigates.

To convert such a trajectory $h$ into a cleaner one aligned with $\tau^\star$, \ours introduces a post-hoc offline trajectory refinement step without re-interacting with the environment. Unlike task refinement, which only has access to partial execution context and must act online, trajectory refinement observes the final task and the entire collected trajectory. This global view allows it to distinguish useful exploratory steps from outdated prefixes, dead-end attempts, or redundant actions that would otherwise become noisy imitation targets. We consider following edits: 
\textbf{Remove}$(i)$ deletes task-irrelevant or redundant steps (e.g., repeated Scroll); 
\textbf{Reorder}$(i,j)$ swaps two locally commutable steps when their targets are independent and have no interfering effects (e.g., open filter then set sort); 
\textbf{Drop}$(h_T)$ discards the entire trajectory if too noisy; and 
\textbf{Keep}$(h_T)$ retains the trajectory as is if already well-aligned.
These edits are proposed by prompting the LLM with $(\tau^\star,h_T)$, where operations are applied conservatively to avoid unintended state changes. The refined pair $(\tau^\star,h_T^\star)$ is less noisy and more suitable for training.
Note that this is an offline post-processing step without environment interaction, thus we implement edits conservatively to avoid unintended state changes.
This design intentionally favors precision over recall: uncertain swaps are rejected rather than risk breaking hidden causal dependencies in the original execution. This behavior is reflected in Table~\ref{tab:reorder_stats}, where reordering is rare, and in Table~\ref{tab:reorder_eval}, where the reordered trajectories receive higher preference and quality scores.

\subsection{Agent Fine-tuning}
After obtaining a high-quality dataset of refined task-trajectory pairs $\mathcal{D} = \{(\tau^{\star}, h^{\star})\}^{N}$, we fine-tune the open-source web agent $f$ to adapt it to such new environment with fully synthetic supervision. We split sample with $T$ steps into a sequence of training examples $\{(\tau^{\star}, \{o_{\leq t}, a_{<t}\}, a_t)\}_{t=1}^{T}$, where the model learns to predict the next action $a_t$ given the task description $\tau^{\star}$ and the historical context of observations and actions $\{o_{\leq t}, a_{<t}\}$. The historical context window is empirically set to $3$ considering training cost and inference latency.
By using the standard supervised fine-tuning (SFT) paradigm, we optimize the model $f$ as follows:

\begin{equation}
  \small
  \mathcal{L}_\text{SFT} = \mathbb{E}_{(\tau^{\star}, h^{\star}) \sim \mathcal{D}} \left[ -\sum_{t=1}^{T} \log p_{\theta}(a_t | \tau^{\star}, o_{\leq t}, a_{<t}) \right] \,.
\end{equation}
After fine-tuning, the agent is expected to better understand the new environment and complete environment specific tasks more effectively, achieving the goal of adaptation.

\section{Experiments}

\subsection{Experimental Setup}
\paragraph{Environment \& Benchmark.}
We conduct experiments in two benchmarks: (1) WebArena~\citep{zhou2024webarenarealisticwebenvironment}, a suite of five websites: e-commerce (Shopping), content management (CMS), social forum (Reddit), developer platform (Gitlab), and map navigation (Maps), which are stable and controllable. (2) Online-Mind2Web~\citep{onlinemind2web}, an online benchmark spanning 136 live websites, which are more realistic and diverse.
The environments provide simplified html code (accessibility tree) and screenshots as observations, and accept common browser interactions listed in Table~\ref{tab:action_table}.
Generally, WebArena is more challenging due to its tasks often requiring authentication, thus involving more complex operations.

% We deliberately choose it as our main testbed for three reasons. 
% First, compared with simplified environments such as WebShop~\citep{yao2022webshop}, each website in WebArena exposes deep task structure (multi pages, authentication, and full functionality), which better matches the realistic scenarios.
% Second, WebArena provides Docker deployments, ensuring the reproducibility, in contrast to online benchmarks such as Mind2Web~\citep{deng2023mind2webgeneralistagentweb} where pages may change over time.
% Finally, since web agents extensively interact with browsers during both exploration and evaluation, running in a local environment keeps experimental cost manageable and avoids the operational risks of live-web interaction.

\paragraph{Baselines \& Models.}
We compare our approach against several strong baselines for synthesis of web agent data, including: (1) Self-Instruct~\citep{wang2023selfinstructaligninglanguagemodels}, which directly generates tasks from a few seed examples via prompting LLMs. (2) OS-Genesis~\citep{sun2025osgenesisautomatingguiagent}, which synthesizes tasks from single-step environment changes with randomly exploring the environment. (3) Explorer~\citep{pahuja2025explorerscalingexplorationdrivenweb}, which synthesizes tasks and trajectories by iteratively refining tasks during trajectory collection.
All baselines are re-implemented using the same LLM of GPT-4.1~\footnote{\href{https://openai.com/index/gpt-4-1/}{https://openai.com/index/gpt-4-1/}} for both task and trajectory synthesis.
We mainly select two popular open-source multimodal LLMs for agent fine-tuning: Qwen2.5-VL-7B-Instruct~\citep{qwen2.5-VL} and UI-TARS-1.5-7B~\citep{uitars}.

\paragraph{Implementation Details.}
For both data synthesis and evaluation, we set a maximum step budget of $30$ per episode, following recommendations from~\citep{zhou2024webarenarealisticwebenvironment}. All methods are synthesizing up to $500$ task-trajectory pairs per website for agent fine-tuning. During execution and training, they are using the same prompt template and action space. We also use the same GPT-4.1 for both synthesis and refinement for \ours. We use a context window of $3$ most recent steps for efficiency.
For agent fine-tuning, we mix synthesized data from five websites to train a single model with a learning rate of 1e-5 and batch size of $32$ for $3$ epochs.
More details are in Appendix~\ref{appendix:implementation}.

\begin{table*}[t]
\centering
\small

% \resizebox{0.9\textwidth}{!}{%

\begin{tabular}{l|l|lllll|l}
    \toprule
    Method & Train on & Shopping & CMS & Reddit & Gitlab & Maps & Overall \\
    \midrule
    GPT-4.1 & - & 30.91 & 24.56 & 15.38 & 26.79 & 21.9 & 25.22 \\
    \midrule
    \midrule
    Qwen  & - & 13.71 & 8.24 & 9.43 & 6.18 & 5.50 & 8.80 \\
    +SFT  & Test Dataset & 27.27 & 12.28 & 19.23 & 10.71 & 12.50 & 16.37 \\
    \midrule
    +Self-Instruct & Synthetic Data & 18.18 & 8.77 & 3.85 & 12.50 & 9.38 & 11.50 \\
    +OS-Genesis   & Synthetic Data & 14.55 & 10.53 & 11.54 & 16.07 & 12.5  & 13.27 \\
    +Explorer     & Synthetic Data & 10.91 & 3.51 & 0.00 & 1.82  & 3.12  & 4.44 \\
    +\textbf{Ours}        & Synthetic Data & \textbf{20.00} & \textbf{21.05} & \textbf{15.38} & \textbf{19.64} & \textbf{28.12} & \textbf{20.80} \\
    \midrule
    \midrule
    UI-TARS & - & 12.73 & 8.77 & 3.85 & 7.14 & 9.38 & 8.85 \\
    +SFT    & Test Dataset & 25.45 & 22.81 & 19.23 & 21.43 & 28.12 & 23.45 \\
    \midrule
    +Self-Instruct & Synthetic Data & 20.00 & 8.77 & 7.69 & 14.55 & 12.50 & 13.33 \\
    +OS-Genesis   & Synthetic Data & \textbf{21.82} & 12.28 & 7.69 & 14.29 & 12.50 & 14.60 \\
    +Explorer     & Synthetic Data & 11.43 & 4.88 & 0.00 & 3.23 & 0.00 & 4.96 \\
    +\textbf{Ours}        & Synthetic Data & 20.00 & \textbf{14.04} & \textbf{19.23} & \textbf{16.07} & \textbf{18.75} & \textbf{17.26} \\
    \bottomrule
    \end{tabular}

% }

\caption{Performance comparison on WebArena benchmark. ``SFT'' means directly use the tasks from test set paired with trajectories collected by GPT-4.1 for agent fine-tuning. The best performance among synthetic methods is highlighted in \text{bold}. In most cases, ours \ours outperforms other methods with synthetic data.}
\label{tab:results}
\end{table*}

\begin{table}[t]
    \centering
    \small
    \resizebox{0.95\columnwidth}{!}{%

    \begin{tabular}{l|ccc|c}
    \toprule
    \diagbox{Method}{Judge} & GPT-4.1 & GPT-5.1 & WebJudge & Avg. \\
    \midrule
    GPT-4.1 & 28.00 & 18.67 & 27.00 & 24.56 \\
    Qwen & 16.90 & 6.49 & 19.48 & 14.29 \\
    \midrule
    Self-Instruct & 17.67 & 13.00 & 19.67 & 16.78 \\
    OS-Genesis & 19.53 & 11.00 & 19.33 & 16.62 \\
    Explorer & 8.33 & 3.67 & 13.33 & 8.44 \\
    \midrule
    \ours & \textbf{31.67} & \textbf{15.67} & \textbf{23.33} & \textbf{23.56} \\
    \bottomrule
    \end{tabular}%

    }

    \caption{Task success rate on the Online-Mind2Web benchmark under three different judges (GPT-4.1, GPT-5.1, and WebJudge~\citep{onlinemind2web}). Our method \ours consistently outperforms baselines across judges, and is competitive with the GPT-4.1.}
    \label{tab:stat_data_mind2web}
\end{table}

\begin{table}[t]
\centering
\small
\resizebox{\columnwidth}{!}{%
\begin{tabular}{l|rrrrr|r}
\toprule
Backbone & Shopping & CMS & Reddit & Gitlab & Maps & Overall \\
\midrule
Qwen3 & 25.45 & 12.28 & 3.85 & 16.07 & 15.62 & 15.93 \\
\ours & \textbf{30.91} & \textbf{17.54} & \textbf{38.46} & \textbf{23.21} & \textbf{15.62} & \textbf{24.34} \\
\bottomrule
\end{tabular}%
}
\caption{Results of Qwen3 on WebArena.}
\label{tab:qwen3_webarena}
\end{table}

\begin{table}[t]
\centering
\small
\resizebox{0.95\columnwidth}{!}{%
\begin{tabular}{l|rrrr}
\toprule
Backbone & GPT-4.1 & GPT-5.1 & WebJudge & Avg. \\
\midrule
Qwen3 & 20.33 & 14.67 & 21.00 & 18.67 \\
\ours & \textbf{28.33} & \textbf{15.00} & \textbf{29.00} & \textbf{24.11} \\
\bottomrule
\end{tabular}%
}
\caption{Results of Qwen3 on Online-Mind2Web.}
\label{tab:qwen3_mind2web}
\end{table}

\subsection{Main Results}
The overall performance of different methods on WebArena is shown in Table~\ref{tab:results}. From the results, we have the following observations:
(1) Training on test set tasks (SFT) provides an informative upper bound for the following experiments using synthetic data.
(2) Self-Instruct struggles without environment grounding, showing limited improvements over base models. This indicates the importance of interacting with the environment for creating environment specific tasks.
(3) Explorer performs surprisingly poorly, even degrading base model performance. This is primarily because its continuous refinement strategy produces overly long trajectories, $68.3\%$ of which exceed the step budget without task completion. This is because continuous task refinement strategy frequently changes task goals and adds new details during execution, causing the agent to deviate from the original objective. Additionally, the original Explorer operates on pages without authentication, limiting its applicability to complex websites, while our re-implementation added the authentication support for fair comparison (details in Appendix~\ref{app:explorer_analysis}).
(4) Our \ours substantially improves web agent adaptation with fully synthetic supervision, consistently outperforming all baselines. Compared to base models and OS-Genesis, we achieve average absolute gains of $+10.2$ and $+5.1$, respectively. \ours also shows closer gap to the upper bound SFT, demonstrating its effectiveness in synthesizing high-quality data.
Additional experiments with Qwen3-VL-8B~\citep{qwen3vl} backbone in Tables~\ref{tab:qwen3_webarena} and~\ref{tab:qwen3_mind2web}. 
Compared with Qwen2.5, Qwen3 starts from stronger base performance, especially on WebArena Shopping and Gitlab. Nevertheless, fine-tuning with the same \ours-synthesized data still improves the overall success rate on both benchmarks. This supports the model-agnostic nature of our pipeline: the refinements operate on the quality of task-trajectory supervision rather than relying on backbone-specific signals. At the same time, the smaller relative gain suggests that part of the behavior captured by synthetic demonstrations may already be internalized by stronger computer-use models, making data quality and task coverage increasingly important as base models improve.

\paragraph{Generalization.}
Beyond the controlled offline WebArena environments with only five websites, we further test the generalization of \ours on Online-Mind2Web, without any synthesis for targeted websites or re-training in Table~\ref{tab:stat_data_mind2web}. We again observe that \ours outperforms other baselines across three different judges and is even competitive with GPT-4.1 though the absolute success rates are modest. These results on both offline (WebArena) and online (Online-Mind2Web) benchmarks suggest that our comprehensive exploration and dual refinement design enables the agent to generalize well on diverse websites rather than overfitting to several fixed environments.

\section{Analysis}
% 1. data analysis
% 2. ablation study
% 3. scaling
% 4. api costs analysis

In this section, we analyze the quality of synthesized data, validate each component's contribution, and examine data scaling of \ours.

\subsection{Synthetic Data Quality}
\begin{table}[t]
    \centering
    \small
    \resizebox{\columnwidth}{!}{

    \begin{tabular}{l|rrr|rr}
    \toprule
    \multicolumn{1}{c|}{\multirow{2}[4]{*}{Method}} & \multicolumn{3}{c|}{Task} & \multicolumn{2}{c}{Trajectory} \\
    \cmidrule{2-6}      & Quality & \# Refine & Diversity & \#Steps & Quality \\
    \midrule
    Self-Instruct & 59.3 (51.8)  & -     & 69 (69)    & 6.3   & 41.7 (45.7)  \\
    OS-Genesis & 56.9 (50.2)  & -     & 65 (76)    & 5.1   & 52.0 (55.6)  \\
    Explorer & 73.1 (72.0)  & 8.6   & 46 (56)    & 20.5  & 36.4 (43.5)  \\
    \midrule
    \ours & 72.6 (71.5)  & 2.0   & 82 (88)    & 7.5   & 82.6 (83.4)  \\
    w/o TR & 68.5 (69.1)  & -     & 84 (85)    & 7.0   & 70.2 (71.9)  \\
    w/o JR & 72.0 (70.4)  & 2.0   & 82 (88)    & 8.8   & 68.7 (75.1)  \\
    \bottomrule
    \end{tabular}%

    }

    \caption{Statistics of synthesized data rated by GPT-5.1 (average across GPT-4.1, GPT-5.1, Qwen3-30B~\citep{yang2025qwen3} in parentheses). Quality is rated on a scale of 0-100. Diversity is measured by 100 samples as a whole. TR and JR stand for task and trajectory refinement. More results are in Appendix~\ref{app:judge_stats}.}
    \label{tab:stat_data}
\end{table}
\begin{table}[t]
    \centering
    \small
    \resizebox{0.95\columnwidth}{!}{

\begin{tabular}{l|ccc|c}
\toprule
Method & \multicolumn{1}{l}{Completed} & \multicolumn{1}{l}{Failed} & \multicolumn{1}{l|}{Exceeded} & Costs \$ \\
\midrule
Explorer & 30.5  & 1.1   & 68.3  & 0.22 \\
\midrule
\ours & 96.5  & 3.5   & 0.0   & 0.13 \\
w/o TR & 47.4  & 49.2  & 3.4   & 0.09 \\
w/o JR & 73.8  & 17.0  & 9.2   & 0.12 \\
\bottomrule
\end{tabular}%

    }

    \caption{Comparison between Explorer and \ours in terms of trajectory completion rates judged by GPT-4.1.  Completed, Failed, and Exceeded are the percentages of trajectories that successfully completed the task, failed to complete the task, or exceeded the maximum step budgets, respectively. Costs are the average API cost per trajectory. }
    \label{tab:stat_complete_rate_traj}
\end{table}
\begin{figure*}
  \centering
  \includegraphics[width=1.0\textwidth]{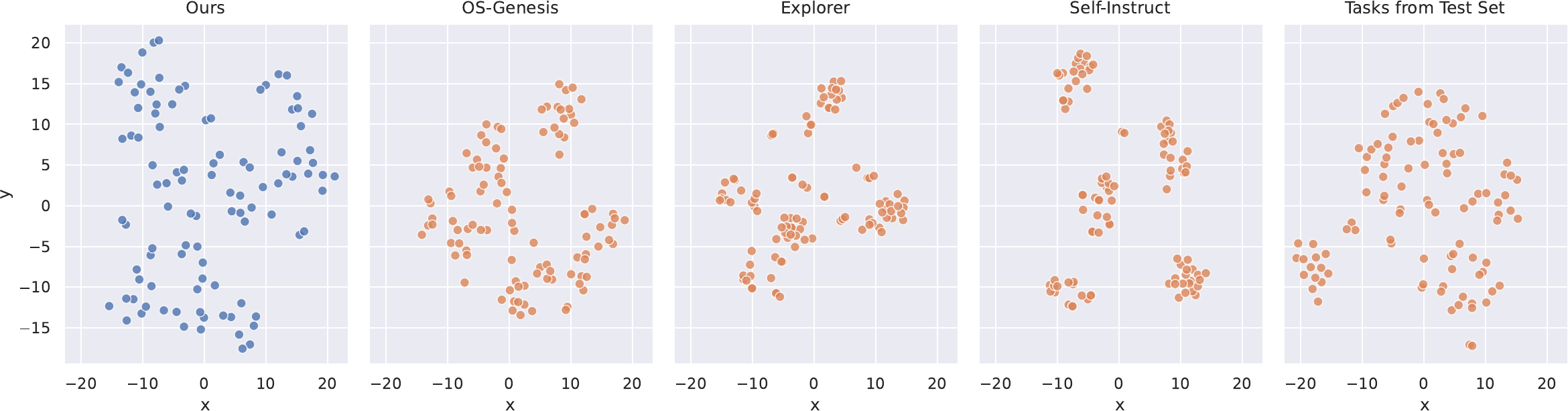}
  \caption{t-SNE visualization of synthesized tasks. Tasks from test set are written by human.}
  \label{fig:tsne}
\end{figure*}

We first analyze the quality and diversity of synthetic data in Table~\ref{tab:stat_data}. To better understand the characteristics of synthesized tasks, we also visualize their embeddings using t-SNE~\citep{tsne} in Figure~\ref{fig:tsne}, where points closer in the plot indicate more similar tasks, while points farther apart indicate greater diversity.

\paragraph{Task Diversity \& Quality.}
Synthesized task diversity is the essential prerequisite for effective agent adaptation.
Self-Instruct produces points heavily clustered in a small region of the embedding space, which indicates, without real environment grounding, LLM ``imagined'' tasks results in narrow task variations. In contrast, OS-Genesis achieves better diversity ($83$) through random environment exploration. Explorer exhibits the poorest diversity ($54$) with a highly clustered pattern, indicating that its initial coarse-grained task proposals from the homepage constrain the space of possible tasks. \ours achieves the highest diversity score after task refinement, with points closely resembling the distribution of human written tasks. This suggests that our categorized exploration and task refinement strategies effectively covers different functions and interaction depths.
To further quantify the coverage behavior of categorized exploration, Table~\ref{tab:cat_stats} reports page-level statistics across environments. The relatively stable category granularity across websites suggests that categorized exploration provides consistent functional coverage despite large differences in page structures. A case study of categorized exploration is provided in Appendix~\ref{app:case_study}.

\begin{table}[t]
  \centering
  \small
  \resizebox{0.95\columnwidth}{!}{%
  \begin{tabular}{l|ccccc}
  \toprule
  Env & Avg. Ele & Avg. Static & Avg. Interact. & Avg. Cat. & Avg. Ele/Cat. \\
  \midrule
  Shopping & 115.5 & 62.3 & 53.3 & 7.9 & 6.7 \\
  CMS & 84.2 & 37.6 & 46.6 & 6.6 & 7.1 \\
  Reddit & 78.6 & 39.7 & 38.9 & 5.3 & 7.3 \\
  Gitlab & 111.2 & 48.9 & 62.3 & 5.8 & 10.7 \\
  Maps & 53.9 & 24.0 & 29.9 & 4.2 & 7.1 \\
  \midrule
  Overall & 88.7 & 42.5 & 46.2 & 6.0 & 7.8 \\
  \bottomrule
  \end{tabular}%
  }
  \caption{Statistics of categorized exploration across all pages visited during task synthesis. ``Avg. Ele'' denotes the average number of DOM elements per page, ``Avg. Static'' and ``Avg. Interact.'' are the averages for non-interactive and interactive elements, respectively, ``Avg. Cat.'' is the average number of functional categories identified by the LLM, and ``Avg. Ele/Cat.'' is the average number of elements per category.}
  \label{tab:cat_stats}
\end{table}

\paragraph{Trajectory Quality.}
Table~\ref{tab:stat_data} shows that our method achieves the highest trajectory quality ($82.6$, avg. $83.4$ across judges), substantially outperforming other methods. The improvement is enabled by our dual refinement: task refinement ensures tasks are feasible and grounded in observations, while trajectory refinement with global context mitigates noise from execution and task edits. 
Notably, our trajectory refinement applies edits conservatively: Table~\ref{tab:reorder_stats} shows that reorder accounts for only $4.1\%$ of edit operations (Remove + Reorder), gated by explicit independence checks. Table~\ref{tab:reorder_eval} further shows that reordered versions achieve higher win rates ($42\%$ vs. $27\%$) and quality scores ($68.9$ vs. $62.1$) in a head-to-head evaluation on 100 reordered trajectories, confirming that constrained reordering improves trajectory quality without introducing instability.

\begin{table}[t]
  \centering
  \small
  \begin{tabular}{l|rrrr}
  \toprule
  Operation & Remove & Reorder & Drop & Keep \\
  \midrule
  Count & 982.5 & 41.8 & 11.0 & 1463.9 \\
  Percentage & 39.3\% & 1.7\% & 0.4\% & 58.6\% \\
  \bottomrule
  \end{tabular}
  \caption{Distribution of trajectory refinement operations. Reorder accounts for only 4.1\% of edit operations (Remove + Reorder), indicating conservative application.}
  \label{tab:reorder_stats}
\end{table}

\begin{table}[t]
  \centering
  \small
  \begin{tabular}{l|cc}
  \toprule
  Metric & Original & Reordered \\
  \midrule
  Win Rate (\%) & 27.0 & 42.0 \\
  Tie Rate (\%) & 31.0 & 31.0 \\
  Quality Score & 62.1 & 68.9 \\
  \bottomrule
  \end{tabular}
  \caption{GPT-5.1 head-to-head evaluation of 100 reordered trajectories. Reordered trajectories achieve higher win rates and quality scores.}
  \label{tab:reorder_eval}
\end{table}

To understand why Explorer underperforms despite similar refinement, we analyze trajectory completion rates in Table~\ref{tab:stat_complete_rate_traj}. Explorer's continuous refinement changes task intentions at each step, causing $68.3\%$ of trajectories to exceed the step budget. In contrast, \ours achieves $96.5\%$ completion rate after trajectory refinement with only $60\%$ of Explorer's OpenAI API cost.

\subsection{Ablation Study}
\begin{table}[t]
    \centering
    \small
    \resizebox{\columnwidth}{!}{

\begin{tabular}{l|rrrrr|r}
\toprule
Method & \multicolumn{1}{c}{Shopping} & \multicolumn{1}{c}{CMS} & \multicolumn{1}{c}{Reddit} & \multicolumn{1}{c}{Gitlab} & \multicolumn{1}{c|}{Maps} & \multicolumn{1}{c}{Overall} \\
\midrule
Qwen  & 13.71 & 8.24  & 9.43  & 6.18  & 5.50  & 8.80 \\
\midrule
\ours & \textbf{20.00} & \textbf{21.05} & \textbf{15.38} & \textbf{19.64} & \textbf{28.12} & \textbf{20.80} \\
w/o CE & 18.18 & 14.04 & \textbf{15.38} & 16.00 & 25.00 & 17.26 \\
w/o TR & 16.37 & 14.04 & \textbf{15.38} & 16.07 & 18.75 & 15.93 \\
w/o JR & 18.18 & 12.28 & 11.54 & \textbf{19.64} & 21.88 & 16.81 \\
w/o TR\&JR & \textbf{20.00} & 14.04 & \textbf{15.38} & 16.07 & 12.50 & 15.93 \\
\bottomrule
\end{tabular}%

    }

    \caption{Ablation study of \ours. CE, TR, and JR stand for categorized exploration, task refinement, and trajectory refinement, respectively.}
    \label{tab:ablation}
\end{table}
To quantify the contribution of each component in \ours, we conduct an ablation study in Table~\ref{tab:ablation}. 
Removing categorized exploration (w/o CE) causes a $3.54\%$ drop, with notable degradation on CMS ($-7.01\%$), confirming that systematic coverage of functional groups is essential for task diversity.
Removing task refinement (w/o TR) leads to a $4.87\%$ drop, the largest among all components. This is because without correcting hallucinations during execution, many synthesized tasks can be infeasible.
Removing trajectory refinement (w/o JR) results in a $3.99\%$ drop, indicating that post-hoc noise removal is crucial for producing clean supervision signals. 
Interestingly, removing both TR and JR (w/o TR\&JR) on Shopping improves while on Maps degrades. This confirms that TR and JR are synergistic: TR introduces some noise to the trajectory due to task edits during execution which JR subsequently removes, and their combination achieves the best overall performance. 

\subsection{Data Scaling}
\begin{figure}[t]
  \centering
  \includegraphics[width=1.0\columnwidth]{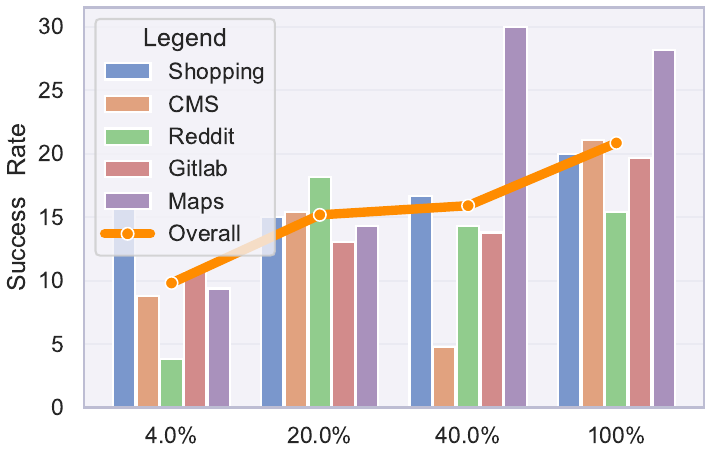}
  \caption{Performance of \ours across different websites with varying synthesis data amounts.}
  \label{fig:data_scaling}
\end{figure}
To evaluate the scalability of \ours, we measure across different data amounts in Figure~\ref{fig:data_scaling}, scaling from $4\%$ (20 tasks per website) to $100\%$ (500 tasks per website). The results demonstrate consistent performance gains as more synthetic data is used, with the average success rate increasing from approximately $10.6$ at $4\%$ data to $20.8$ at $100\%$ data. Notably, different websites exhibit varying performance gains, which may be attributed to their inherent task complexity and capacity. For example, Map environment reaches the peak at $40\%$ data, while CMS and Gitlab continue to steadily improve.
The overall scaling behavior shows that \ours maintains data quality consistently as the synthesis scales with the website complexity.

\section{Conclusion}
In this paper, we study how to adapt web agents to new environments where environment-specific tasks and demonstrations are scarce.
We identify that existing synthetic data generation methods suffer from severe quality issues: tasks often contain hallucinations and collected trajectories are noisy. To address these issues, we propose \ours, a fully synthetic supervision framework that improves data quality through dual refinement of both tasks and trajectories. In addition, we introduce categorized exploration to systematically cover web elements and interactions, enhancing the diversity and efficiency of task synthesis. 
Extensive experiments and analysis demonstrate the effectiveness of our approach in adapting web agents to new environments.
Beyond adaptation, we believe \ours is a valuable resource for the rapidly emerging field of agentic reinforcement learning where large-scale and diverse tasks are crucial.

\clearpage
\newpage
\section*{Limitations}
While this paper demonstrates promising results for adapting web agents, we acknowledge several limitations that may help future research.
\begin{enumerate}[label=(\arabic*), itemsep=0pt]
  \item We evaluate on both offline (WebArena~\citep{zhou2024webarenarealisticwebenvironment}) and online (Online-Mind2Web~\citep{onlinemind2web}) benchmarks. WebArena provides realistic complexity with multi-page interactions, authentication, and full functionality, while Online-Mind2Web spans 136 live websites for generalization testing. Other offline benchmarks such as Mind2Web~\citep{deng2023mind2webgeneralistagentweb} and WebShop~\citep{yao2022webshop} provide only static snapshots or simplified environments, preventing the interactive exploration that \ours relies on for data synthesis. Future work may explore deploying data synthesis on live websites, though this presents significant challenges: live sites constantly change and often require CAPTCHA solving, extensive automated visits may trigger security mechanisms, and synthesized data from live websites cannot be easily released due to copyright and privacy constraints, limiting reproducibility.
  \item For task and trajectory synthesis, we rely solely on GPT-4.1 without hyperparameter tuning though we provide quality assessments using different LLMs. We do not optimize agent execution parameters such as the maximum execution steps, exploration depth, or sampling strategies during data collection. Additionally, we do not explore alternative prompting strategies or more advanced LLMs that may further improve synthesis quality. 
  \item Our agent fine-tuning employs standard SFT on the synthetic data. We do not explore more advanced training algorithms such as Direct Preference Optimization~\citep{rafailov2023direct}, or online reinforcement learning methods that could potentially yield better performance. Also, incorporating richer supervision signals (e.g., inner monologue annotations, structured reasoning traces) during training may lead to substantially larger performance gains. 
  However, these advanced approaches typically require constructing verification mechanisms or reward models, which is beyond the scope of this paper.

\end{enumerate}

\section*{Ethics Statement}
Our research is conducted entirely on controlled, self-hosted environments. However, the proposed method could potentially be misused on live websites to enable unauthorized data scraping, or account related security issues. We strongly discourage such misuse and emphasize that all deployment of web agents must comply with applicable laws, website terms of service, and ethical guidelines. Future work should incorporate appropriate safeguards and detection of sensitive information. We are committed to responsible research and encourage the community to develop autonomous web agents with consideration for security, privacy, and ethical implications.

\section*{Use of AI Assistants}
We acknowledge the use of AI Assistants in the writing of this manuscript, mainly for correcting grammatical errors and improving sentence fluency. All the generated content was carefully reviewed by the authors.

\bibliography{custom}

\appendix
\newpage
\section{Implementation Details} 
\begin{figure*}[h]
  \centering
  \includegraphics[width=1.0\textwidth]{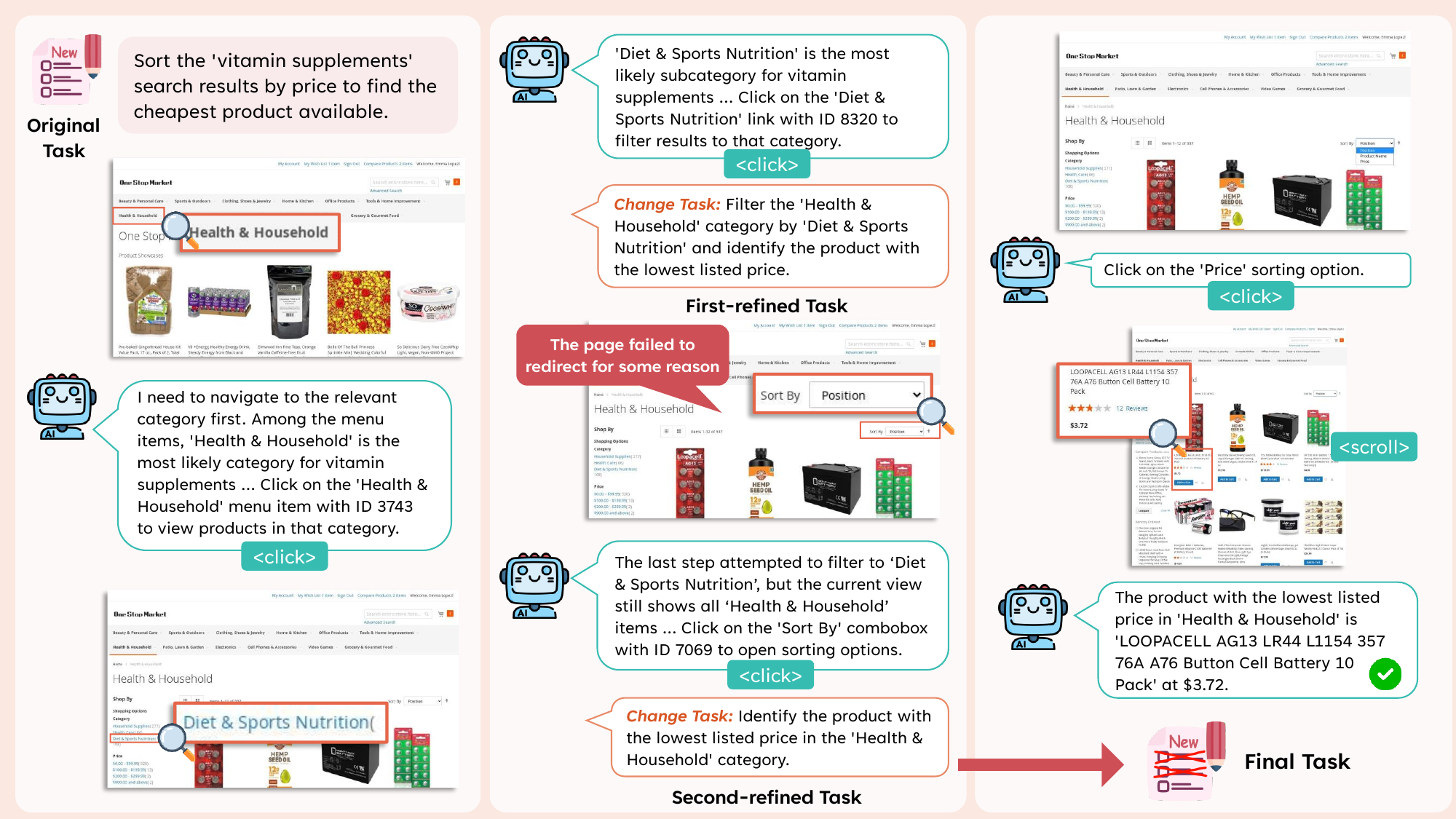}
  \caption{Case study of Task Refinement. During task execution, \ours detects when a task becomes infeasible based on the progression of its execution and refines the task accordingly, guided by the observed failure and current environment state.}
  \label{fig:task-refine}
\end{figure*}

\begin{figure*}[h]
  \centering
  \includegraphics[width=1.0\textwidth]{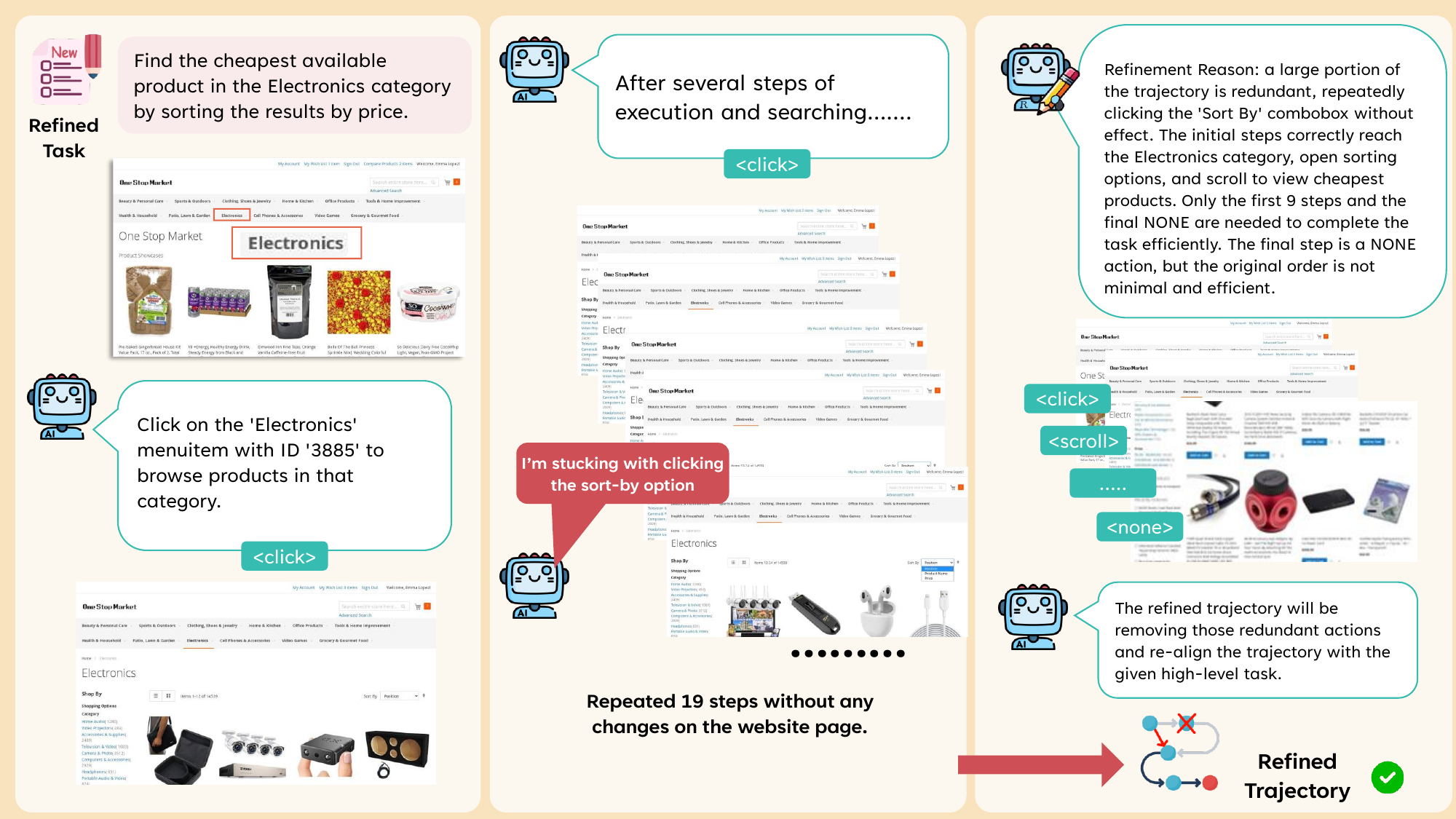}
  \caption{Case study of Trajectory Refinement. After trajectory collection, \ours evaluates the trajectory and refines it by reordering steps or removing unnecessary actions to ensure the trajectory consistently align with the task objective and ends with a successful completion.}
  \label{fig:case-trajectory-refine}
\end{figure*}

\subsection{Environment}
\label{appendix:implementation}

\begin{table}[h]
\centering
\small
\begin{tabularx}{\columnwidth}{@{}lX@{}}
\toprule
Action & Description \\
\midrule
\texttt{click [id]} & Clicks on an element with the given \texttt{element\_id}. \\[2pt]
\texttt{type [id] [text]} & Types the specified text into the field with \texttt{element\_id}. \\[2pt]
\texttt{hover [id]} & Moves the cursor over an element with the given \texttt{element\_id}. \\[2pt]
\texttt{press [key\_comb]} & Presses a keyboard shortcut (e.g., \texttt{Ctrl+V}, \texttt{Cmd+V}). \\[2pt]
\texttt{scroll [up|down]} & Scrolls vertically in the given direction. \\[2pt]
\texttt{goto [url]} & Navigates directly to the specified URL. \\[2pt]
\texttt{go\_back} & Navigates to the previous page in browser history. \\[2pt]
\texttt{go\_forward} & Navigates to the next page in browser history. \\[2pt]
\texttt{none [summary]} & Produces a summary or final answer without any browser action. \\[2pt]
\texttt{stop [reason]} & Stops execution when the task is impossible or inconsistent. \\
\bottomrule
\end{tabularx}
\caption{Available atomic low-level browser interactions and their descriptions for WebArena.}
\label{tab:action_table}
\end{table}

\paragraph{Action Space.} All the actions that are executable in the trajectories are shown in Table~\ref{tab:action_table} with regular web actions. The ``none'' action signals the final step for answering the task. For example, if the task is to find the cheapest product on a shopping website, the last action would be ``none'' with the summary of the cheapest product as the observation. The ``stop'' action is used to terminate the trajectory when the agent exceeds the maximum step budget or encounters an unrecoverable error.

\paragraph{Evaluation.}
WebArena~\citep{zhou2024webarenarealisticwebenvironment} contains 812 challenging web navigation tasks derived from 241 human written task templates, including maps, e-commerce, Reddit forums, and software development. Due to the naturally high cost of time interacting with the browser, we follow prior work~\citep{sun2025osgenesisautomatingguiagent} and use 226 tasks for evaluation, where we include only one task per template and exclude tasks that require cross-website navigation. The numbers of tasks for each website are 55, 57, 26, 56, and 32 for Shopping, CMS, Reddit, Gitlab, and Maps, respectively. Note that for training with tasks from test set (SFT method in Table~\ref{tab:results}), we still use the full set of 812 tasks to enrich the training data as much as possible.

Online-Mind2Web~\citep{onlinemind2web} contains 300 realistic and diverse tasks spanning 136 live websites. Each task is requiring the agent to interact with real websites. The evaluation is performed with LLM-as-a-Judge through key point identification, key screenshot identification, and final outcome judgment.

\begin{table}[t]
  \centering
  \small
  \resizebox{\columnwidth}{!}{
  \begin{tabular}{l|rrr|rr}
  \toprule
  \multicolumn{1}{c|}{\multirow{2}[4]{*}{Method}} & \multicolumn{3}{c|}{Task} & \multicolumn{2}{c}{Trajectory} \\
  \cmidrule{2-6}      & Quality & \# Refine & Diversity & \#Steps & Quality \\
  \midrule
  Self-Instruct & 49.2  & -     & 69    & 6.3   & 56.6  \\
  OS-Genesis & 39.8  & -     & 83    & 5.1   & 54.1  \\
  Explorer & 69.8  & 8.6   & 54    & 20.5  & 48.1  \\
  \midrule
  \ours & 70.2  & 2.0   & 95    & 7.5   & 92.5  \\
  w/o TR & 68.2  & -     & 86    & 7.0   & 78.1  \\
  w/o JR & 69.5  & 2.0   & 95    & 8.8   & 86.3  \\
  \bottomrule
  \end{tabular}%
  }
  \caption{Trajectory quality rated by GPT-4.1.}
  \label{tab:stat_data_gpt4}
\end{table}

\begin{table}[t]
  \centering
  \small
  \resizebox{\columnwidth}{!}{
  \begin{tabular}{l|rrr|rr}
  \toprule
  \multicolumn{1}{c|}{\multirow{2}[4]{*}{Method}} & \multicolumn{3}{c|}{Task} & \multicolumn{2}{c}{Trajectory} \\
  \cmidrule{2-6}      & Quality & \# Refine & Diversity & \#Steps & Quality \\
  \midrule
  Self-Instruct & 46.8  & -     & 68    & 6.3   & 38.9  \\
  OS-Genesis & 53.9  & -     & 80    & 5.1   & 60.7  \\
  Explorer & 73.0  & 8.6   & 68    & 20.5  & 45.9  \\
  \midrule
  \ours & 71.8  & 2.0   & 88    & 7.5   & 75.0  \\
  w/o TR & 70.5  & -     & 85    & 7.0   & 67.5  \\
  w/o JR & 69.8  & 2.0   & 88    & 8.8   & 70.3  \\
  \bottomrule
  \end{tabular}%
  }
  \caption{Trajectory quality rated by Qwen3-30B.}
  \label{tab:stat_data_qwen}
\end{table}

\begin{table*}[t]
    \centering
    \setlength{\fboxrule}{0.8pt}
    \fbox{%
        \begin{minipage}{0.97\textwidth}
        \small
        \setlength{\parskip}{2pt}
        \vspace{2pt}
        % \texttt{\textbf{Prompt for Web GUI Element Grouping and Categorization}}\\[4pt]
        % \texttt{prompt = \{ }\\[3pt]
        % \texttt{``intro'': \texttt{``}\texttt{``}\texttt{``}
        \texttt{You are a GUI (Graphical User Interface) Web Agent expert capable of grouping interactive elements from a web page into high-level user task categories.}\\[6pt]

        \texttt{**Information**}\\[3pt]
        \texttt{Current Page:}\\
        \texttt{- URL:}\\
        \texttt{\{url\}}\\
        \texttt{- Accessibility Tree (only current view, not full page):}\\
        \texttt{\{page\_context\}}\\
        \texttt{- Elements:}\\
        \texttt{\{elements\}}\\
        \texttt{- Screenshot of the page (only current view, not full page):}\\
        \texttt{<image is provided in the attachment>}\\[6pt]

        \texttt{**Your Goal**}\\[3pt]
        \texttt{1. Fully explore the current page and its content to understand its functionality and layout.}\\
        \texttt{2. Categorize ALL provided \{element\_num\} elements into different categories (list[dict]) based on their natural purpose.}\\
        \texttt{3. Add a \{const\_uninteractive\_category\} category (list[int]) for non-interactive elements that cannot be CLICK, TYPE, or HOVER.}\\
        \texttt{4. For each category (except \{const\_uninteractive\_category\}), decide:}\\
        \texttt{\{ "action": choose from [CLICK, TYPE, HOVER], "element\_id": id\_of\_element (int), "value": if TYPE, provide text to type; else '', "low-level\_instruction": concise description of the action \}}\\[3pt]
        \texttt{Example low-level instructions:}\\
        \texttt{- "Click on the 'Add to Cart' button next to the product to add it to your shopping cart."}\\
        \texttt{- "Type 'OpenAI' into the search bar to find relevant articles."}\\
        \texttt{- "Scroll down to view the latest blog posts on the homepage."}\\[4pt]

        \texttt{5. Provide an appropriate and meaningful value for "value" if the action is TYPE. Examples:}\\
        \texttt{- For a search box, generate a realistic search query.}\\
        \texttt{- For a textbox, generate plausible text according to context.}\\[6pt]

        \texttt{**Output Requirements**}\\[3pt]
        \texttt{Return ONLY a JSON dictionary (no commentary) with the following format:}\\
        \texttt{\{ }\\
        \texttt{~~"Analysis": "your analysis of the current page state and elements",}\\
        \texttt{~~"Categorization": \{ }\\
        \texttt{~~~~"category\_1": [}\\
        \texttt{~~~~~~\{ "action": "xxx", "element\_id": int, "value": "xxx or empty string", "low-level\_instruction": "a concise instruction" \}, ...other actions... ],}\\
        \texttt{~~~~"category\_2": [...],}\\
        \texttt{~~~~"category\_3": [...],}\\
        \texttt{~~~~...\{different categories\}...,}\\
        \texttt{~~~~"\{const\_uninteractive\_category\}": [element\_id\_1, element\_id\_2, ...]}\\
        \texttt{~~\}}\\
        \texttt{\}}\\[3pt]
        \texttt{RETURN ME THE DICTIONARY I ASKED FOR WITHOUT ANY COMMENTARY.}\\[2pt]
        \texttt{"""}\\[2pt]
        \vspace{1pt}
        \end{minipage}
    }
    \caption{Prompt for web elements categorization.}
    \label{tab:cat-prompt}
\end{table*}

\begin{table*}[t]
    \centering
    \setlength{\fboxrule}{0.8pt}
    \fbox{%
        \begin{minipage}{0.97\textwidth}
        \scriptsize
        \setlength{\parskip}{1pt}
        \vspace{1pt}
        % \texttt{\textbf{Prompt for OS-Genesis for Tasks generation}}\\[4pt]
        % \texttt{prompt = \{ }\\[3pt]
        % \texttt{``intro'': \texttt{``}\texttt{``}\texttt{``}
        \texttt{You are a GUI (Graphical User Interface) expert capable of analyzing interface changes and envisioning executable tasks or instructions. Given a GUI interface change caused by an action (e.g., clicking or typing) and the corresponding element highlighted in red boxes, you are required to analyze the interface and generate related tasks.}\\[3pt]

        \texttt{Your task is to envision tasks based on the current action and the resulting changes in the screenshots. The output should include three components:}\\[2pt]

        \texttt{1. Sub-Instruction: Create a natural language instruction for the current action based on the interface changes it caused. The instruction should be concise, clear, and actionable, incorporating specific details critical to the task, such as elements, file names, timestamps, or other relevant content visible in the screenshots. For example:}\\
        \texttt{- "Click on the 'Add to Cart' button next to the product to add it to your shopping cart."}\\
        \texttt{- "Type 'OpenAI' into the search bar to find relevant articles."}\\
        \texttt{- "Scroll down to view the latest blog posts on the homepage."}\\[2pt]

        \texttt{2. Analysis: Carefully analyze the before-and-after screenshots step by step, focusing on the changes caused by the action. Then, examine key elements in both screenshots and consider possible operations based on these elements. For example: "The previous screen displayed the main interface of a shopping website, featuring multiple product categories and several showcased items. After clicking the 'Sign Up' button, the interface transitioned to a login page where an email and password can be entered to log into an account. The login page also provides other options, such as recovering a password, creating a new account, or logging in with a Google account."}\\[2pt]

        \texttt{3. High-Level Instruction: Based on the before-and-after screenshots, the action, and the analysis, generate a high-level task that you believe can be completed within the current interface. There are three types of tasks:}\\
        \texttt{- Information seeking: The user wants to obtain certain information from the webpage, such as product details, reviews, map information, or route comparisons. Please propose clear and specific questions that need an explicit answer, and avoid asking for summary-type questions, such as "summarize the information about a product."}\\
        \texttt{- Site navigation: The user wants to navigate to a specific page or state.}\\
        \texttt{- Content modification: The user wants to modify the content of a webpage or its settings.}\\[2pt]

        \texttt{The high-level instruction should be creative. You need to deeply analyze the elements and executable actions on the interface to generate realistic, valuable, and executable tasks that can be completed within the current GUI. The instruction should be specific, actionable, and goal-oriented, ensuring the task can be completed on the current GUI by including all critical specifics such as file names, relevant timings, or required details.}\\[3pt]

        \texttt{Below is a brief description of the current website: \{website\_intro\}}\\
        \texttt{Here are some examples of High-Level Instruction for reference: \{task\_examples\}}\\[2pt]

        \texttt{Current Action: \{current\_action\_str\}}\\
        \texttt{Website Name: \{website\_name\}}\\[2pt]

        \texttt{Before-action Screenshot: <image is provided in the first attachment> (the action's target element is highlighted in red box if applicable)}\\
        \texttt{After-action Screenshot: <image is provided in the second attachment>}\\[2pt]

        \texttt{Please generate tasks that can be completed on the current platform, and avoid tasks that are unrelated to the current website.}\\[3pt]

        \texttt{You ONLY need to return a JSON dictionary formatted as follows (no extra commentary):}\\
        \texttt{\{ }\\
        \texttt{~~"Sub-Instruction": "xxx",}\\
        \texttt{~~"Analysis": "xxx",}\\
        \texttt{~~"High-Level-Instruction": "xxx"}\\
        \texttt{\}}\\[2pt]
        \texttt{RETURN ONLY THE DICTIONARY I ASKED FOR.}\\[1pt]
        \texttt{\}}\\[1pt]
        \vspace{0pt}
        \end{minipage}
    }
    \caption{Prompt for agentic task generation, adopted from OS-Genesis~\citep{sun2025osgenesisautomatingguiagent}.}
    \label{tab:webarena-zero-shot-eval-prompt}
\end{table*}

\begin{table*}[t]
    \centering
    \setlength{\fboxrule}{0.8pt}
    \fbox{%
        \begin{minipage}{0.97\textwidth}
        \scriptsize
        \setlength{\parskip}{1pt}
        \vspace{1pt}
        % \texttt{\textbf{Prompt for Long-Horizon Web GUI Planning (Next Action)}}\\[4pt]
        % \texttt{prompt = \{ }\\[3pt]
        % \texttt{``intro'': \texttt{``}\texttt{``}\texttt{``}
        \texttt{You are a GUI (Graphical User Interface) Web Agent expert capable of long-horizon planning and executing high-level tasks on a website. Based on the observations and the high-level task to complete, generate the next low-level instruction.}\\[3pt]

        \texttt{**Information**}\\[2pt]
        \texttt{1. High-Level Task (your ultimate goal to finish):}\\
        \texttt{"\{high\_level\_task\}"}\\[2pt]

        \texttt{2. Current Page (only current view, not full page, you may need to scroll to see more):}\\
        \texttt{- URL:}\\
        \texttt{\{url\}}\\
        \texttt{- Accessibility Tree (Page Context):}\\
        \texttt{\{page\_context\}}\\
        \texttt{- Elements (addressable in this view):}\\
        \texttt{\{elements\}}\\
        \texttt{- Screenshot (only current view, not full page):}\\
        \texttt{\{img\_info\}}\\[2pt]

        \texttt{3. History of Actions (\{hint\_for\_history\}):}\\
        \texttt{\{previous\_state\_action\}}\\[3pt]

        \texttt{**Critical Rules for Success**}\\[2pt]
        \texttt{1. Issue only actions valid for the current observation (elements, accessibility tree, screenshot).}\\
        \texttt{2. Propose ONE atomic action per item in your Potential-Actions list; actions must be independently executable.}\\
        \texttt{3. Prefer element IDs from the current Elements list for CLICK/TYPE/HOVER.}\\
        \texttt{4. Provide meaningful non-empty value if action $\in$ \{TYPE, SCROLL, GOTO, NONE, STOP\}.}\\
        \texttt{5. If the task is complete, use NONE with the final answer in value; do not propose further actions.}\\
        \texttt{6. Be concise, avoid redundant/risky actions; each action must advance the task.}\\
        \texttt{7. If the task is hallucinated/low-quality/impossible, cautiously choose STOP based on observations/history.}\\[2pt]

        \texttt{Pseudo-code for deciding STOP:}\\
        \texttt{if high\_level\_task lacks required info → STOP}\\
        \texttt{if high\_level\_task contains hallucinations → STOP}\\
        \texttt{if task is inappropriate/harmful → STOP}\\
        \texttt{if multiple ($\geq$3) similar attempts already failed → STOP}\\
        \texttt{else → consider NON-STOP actions}\\[2pt]

        \texttt{8. First write a "state\_observation\_summary", then do step-by-step "reasoning", then decide "next\_action".}\\
        \texttt{9. Expect MULTIPLE steps; choose the next action that changes state; continue iteratively.}\\
        \texttt{10. You MUST actively decide the next step; do not choose NONE/STOP unless certain of finish/impossibility.}\\
        \texttt{11. In "reasoning", explicitly apply the STOP vs NON-STOP pseudo-code.}\\
        \texttt{12. Actively explore alternatives before STOP if current approach stalls.}\\
        \texttt{13. Choose elements strictly from "Elements (addressable in this view)"; justify this choice in "reasoning".}\\
        \texttt{14. If the page doesn't change after an action, consider SCROLL to reveal more elements.}\\
        \texttt{15. Special note: when typing a date, use "MM/DD/YYYY".}\\[3pt]

        \texttt{**Output Requirements**}\\[2pt]
        \texttt{Return ONLY a JSON dictionary (no commentary) with:}\\
        \texttt{\{ }\\
        \texttt{~~"state\_observation\_summary": "1--3 sentence summary of the current state relevant to the task",}\\
        \texttt{~~"reasoning": "step-by-step reasoning to decide the next action; include rule-based justification and STOP check",}\\
        \texttt{~~"next\_action": \{ }\\
        \texttt{~~~~"low-level\_instruction": "concise string, e.g., click the button with ID 'submit'",}\\
        \texttt{~~~~"action": \{ "type": "XXXX", "element\_id": <int or '' (if int, it MUST exist in Elements)>, "value": <string or '' per spec> \}}\\
        \texttt{~~\}}\\
        % \texttt{\}}\\[3pt]
        % \texttt{\}}\\[2pt]
        \vspace{0pt}
        \end{minipage}
    }
    \caption{Prompt for long-horizon planning and next-action generation of web agent.}
    \label{tab:cot-prompt}
\end{table*}

\begin{table*}[t]
    \centering
    \setlength{\fboxrule}{0.8pt}
    \renewcommand{\baselinestretch}{0.05} % adjust overall line spacing
    \fbox{%
        \begin{minipage}{0.97\textwidth}
        \small
        \vspace{2pt}
        % \texttt{\textbf{Prompt for Associating High-Level Tasks}}\\[4pt]
        \texttt{You are a GUI (Graphical User Interface) Web Agent expert specializing in analyzing interface changes and determining whether a high-level task should be refined based on a series of agent actions and observations.}\\[4pt]

        \texttt{\#\# High-Level Task Categories}\\[3pt]
        \texttt{1. **Information Seeking** — User aims to retrieve specific information from the website.}\\
        \texttt{- Examples:}\\
        \texttt{~~- "What is the most expensive product in the 'Electronics' category?"}\\
        \texttt{~~- "What are the top 5 posts in the 'Technology' forum?"}\\
        \texttt{~~- "Summarize the reviews for the product 'iPhone 11'."}\\[4pt]

        \texttt{2. **Site Navigation** — User aims to reach a specific page or site state.}\\
        \texttt{- Examples:}\\
        \texttt{~~- "Go to the billing page to check the latest transactions."}\\
        \texttt{~~- "Navigate to the 'Contact Us' page and fill out the form to express interest in joining the company."}\\
        \texttt{~~- "Find the wiki page of 'the youngest person to receive a Nobel Prize'."}\\[4pt]

        \texttt{3. **Content Modification** — User aims to change site content or settings.}\\
        \texttt{- Examples:}\\
        \texttt{~~- "Create a user account with username 'bob2134' and password '128nxc18zxv'."}\\
        \texttt{~~- "Post a new article titled 'The Future of AI' in the 'Technology' forum."}\\
        \texttt{~~- "Create a code repo named 'Agent' and add a README with the text 'This is a code repo for an intelligent agent.'"}\\[5pt]

        % \texttt{---}\\[3pt]

        \texttt{\#\# Refine Rules}\\[3pt]
        \texttt{\#\#\# When to REFINE the task}\\
        \texttt{Refine the task if the following situations are observed (cite triggers in "Analysis"):}\\
        \texttt{1. **Invalid or Inconsistent Goal** — target entity/page/action does not exist, cannot be located, or conflicts with observed facts.}\\
        \texttt{2. **Insufficient Executable Details** — essential parameters are missing and cannot be inferred.}\\
        \texttt{3. **Stalled or Repetitive Execution** — three or more consecutive actions show no meaningful change, or same error repeats.}\\[3pt]

        \texttt{\#\#\# When NOT to REFINE}\\
        \texttt{- Goal is valid and consistent with observations.}\\
        \texttt{- Essential parameters are available or can be inferred.}\\
        \texttt{- Actions show measurable progress.}\\
        \texttt{- No persistent or repetitive failures detected.}\\[3pt]

        \texttt{\#\#\# How to REFINE}\\
        \texttt{If refinement is required (your analysis must reference the below rules):}\\
        \texttt{1. **Concretize Missing Details** — add essential parameters from history or observation.}\\
        \texttt{2. **Align with Reality** — replace hallucinated entities with actual ones found on the site.}\\
        \texttt{3. **Downscope the Goal** — adjust to the next achievable milestone.}\\
        \texttt{4. **Preserve Task Type** — keep within same category unless required otherwise.}\\

        % \texttt{---}\\[3pt]
        \texttt{\#\# Goal}\\
        \texttt{Ensure the refined task is either already completed or highly likely to complete within the next 1–2 steps.}\\[4pt]

        % \texttt{---}\\[3pt]
        \texttt{\#\# Output Requirements}\\
        \texttt{- Format: JSON dictionary only, no commentary.}\\
        \texttt{- Fields:}\\
        \texttt{~~- "Analysis": Step-by-step reasoning.}\\
        \texttt{~~- "Need-to-Refine": "yes" or "no".}\\
        \texttt{~~- "High-Level-Task": Refined task if "yes", else empty string.}\\[4pt]

        % \texttt{---}\\[3pt]
        \texttt{\#\# Information}\\
        \texttt{1. Current High-Level-Task: "\{current\_high\_level\_task\}"}\\
        \texttt{2. Previous High-Level-Tasks (oldest to newest):}\\
        \texttt{<start\_previous\_high\_level\_tasks>}\\
        \texttt{\{previous\_high\_level\_tasks\}}\\
        \texttt{<end\_previous\_high\_level\_tasks>}\\[3pt]

        \texttt{3. History of Actions (\{hint\_for\_history\}):}\\
        \texttt{<start\_action>}\\
        \texttt{\{previous\_state\_action\}}\\
        \texttt{<end\_action>}\\[3pt]

        \texttt{4. Current Page (only current view):}\\
        \texttt{- URL: "\{curr\_url\}"}\\
        \texttt{- Page Context:}\\
        \texttt{<start\_context>}\\
        \texttt{\{curr\_state\_context\}}\\
        \texttt{<end\_context>}\\
        \texttt{- Screenshot: "\{img\_info\}"}\\[4pt]

        \texttt{---}\\[3pt]
        \texttt{You ONLY need to return a JSON dictionary formatted as follows (no commentary):}\\
        \texttt{\{ }\\
        \texttt{~~"Analysis": "step-by-step reasoning",}\\
        \texttt{~~"Need-to-Refine": "yes or no",}\\
        \texttt{~~"High-Level-Task": "refined task if yes, otherwise empty"}\\
        \texttt{\}}\\[2pt]
        \texttt{RETURN ONLY THE DICTIONARY I ASKED FOR.}\\[2pt]
        % \texttt{\}}\\
        \vspace{1pt}
        \end{minipage}
    }
    \caption{Prompt for refining tasks during trajectory collection.}
    \label{tab:association-web-prompt}
\end{table*}

\begin{table*}[t]
    \centering
    \setlength{\fboxrule}{0.8pt}
    \fbox{%
        \begin{minipage}{0.97\textwidth}
        \small
        \vspace{2pt}

        \texttt{You are a GUI (Graphical User Interface) Web Agent expert. Your job is to analyze a high-level task and its trajectory (sequence of states and actions), assign a quality score, and decide one of:}\\
        \texttt{- "keep": keep the trajectory as-is (already minimal, ordered, and ends with a correct NONE action with a non-empty value).}\\
        \texttt{- "refine": reorder or delete steps to make the trajectory succeed (final step must be NONE with a non-empty explanation).}\\
        \texttt{- "drop": discard the trajectory entirely (e.g., irreparable, hallucinatory, impossible, unsafe, or missing critical information).}\\[6pt]

        \texttt{A trajectory is structured as:}\\
        \texttt{"Length of trajectory, High-level task, summary of state1, action1, summary of state2, action2, ..."}\\[6pt]

        \texttt{\#\#\# Scoring Rubric (0--100)}\\
        \texttt{Evaluate the trajectory on:}\\
        \texttt{1. Goal Alignment (0--25): Steps relevant to the high-level task.}\\
        \texttt{2. Logical Order (0--25): Steps follow a coherent and sensible sequence.}\\
        \texttt{3. Efficiency (0--25): Avoids redundant or unnecessary actions.}\\
        \texttt{4. Success Likelihood (0--25): Likely to end successfully with NONE (non-empty value).}\\[4pt]

        \texttt{Note: The score is advisory; the final decision (keep/refine/drop) depends on qualitative judgment.}\\[6pt]

        \texttt{---}\\[3pt]
        \texttt{\#\#\# Decision Policy}\\
        \texttt{- Always ensure kept/refined trajectories end with a NONE action and non-empty value.}\\
        \texttt{- If refining, reorder or delete existing steps (do not add new ones).}\\
        \texttt{- Replace STOP with NONE if success is achievable.}\\
        \texttt{- If dropping, do not fabricate NONE; instead, provide a clear drop\_reason.}\\[4pt]

        \texttt{**Indexing and Deletion Rules:**}\\
        \texttt{- Let the trajectory contain K (observation, action) pairs, indexed 0..K-1.}\\
        \texttt{- Reordering: return indices in a new order.}\\
        \texttt{- Deletion: omit indices.}\\
        \texttt{- No duplicates or out-of-range indices. Do not invent new steps.}\\[6pt]

        \texttt{\#\#\# Input}\\
        \texttt{\{trajectory\}}\\[6pt]

        \texttt{\#\#\# Output Requirement (STRICT)}\\
        \texttt{Return ONLY one JSON object (no extra text, no code fences):}\\
        \texttt{\{}\\
        \texttt{~~"task": "<exact high-level task string>",}\\
        \texttt{~~"score": <int>,~~// 0--100, advisory only}\\
        \texttt{~~"decision": "keep" | "refine" | "drop",}\\
        \texttt{~~"order": [<int>, ...],~~// indices in final order}\\
        \texttt{~~"modify\_end": <true|false>,}\\
        \texttt{~~"append\_end": <true|false>,}\\
        \texttt{~~"final\_none\_value": "<non-empty explanation for final NONE>",}\\
        \texttt{~~"drop\_reason": "<reason if dropped>",}\\
        \texttt{~~"modification\_reason": "<brief rationale for keep/refine/drop>"}\\
        \texttt{\}}\\[3pt]

        \texttt{\#\#\# Additional Constraints}\\
        \texttt{- "task" must match the high-level task exactly.}\\
        \texttt{- If decision = keep: order = [0,1,...,K-1], and final step already NONE.}\\
        \texttt{- If decision = refine: must end with NONE and valid order.}\\
        \texttt{- If decision = drop: order = [], final\_none\_value empty, provide drop\_reason.}\\[2pt]

        \vspace{1pt}
        \end{minipage}
    }
    \caption{Prompt for trajectory quality evaluation and refinement.}
    \label{tab:y-refine-prompt}
\end{table*}

\begin{table*}[t]
    \centering
    \setlength{\fboxrule}{0.8pt}
    \renewcommand{\baselinestretch}{0.05} % adjust overall line spacing
    \fbox{%
        \begin{minipage}{0.97\textwidth}
        \small
        \vspace{2pt}

        \texttt{You are a GUI (Graphical User Interface) Web Agent evaluation expert. Your job is to evaluate the **diversity** of a set of high-level tasks generated for a single web environment.}\\[3pt]
        \texttt{Each task represents a distinct user goal that can be accomplished on the same webpage (e.g., "Search for a product", "Sort results by price", "View product details"). Your evaluation should judge how **broad, non-overlapping, and complementary** these tasks are relative to one another.}\\[5pt]

        \texttt{---}\\[3pt]
        \texttt{\#\# Input}\\
        \texttt{The input contains multiple high-level tasks that share the same web environment:}\\
        \texttt{\{task\_list\_block\}}\\[5pt]

        \texttt{---}\\[3pt]
        \texttt{\#\# Scoring Rubric (0--100)}\\
        \texttt{Evaluate the **diversity** of the provided task set using the following criteria:}\\[2pt]
        \texttt{1. **Intent Variety (0--25):** Do the tasks represent different user intents (e.g., information seeking vs. navigation vs. modification)?}\\
        \texttt{2. **Action Diversity (0--25):** Do the tasks require different types of GUI interactions (e.g., clicking, typing, scrolling, submitting forms)?}\\
        \texttt{3. **Goal Coverage (0--25):** Do the tasks explore different meaningful aspects or functionalities of the environment?}\\
        \texttt{4. **Redundancy Minimization (0--25):** Are there minimal duplicate or near-duplicate tasks (i.e., no rephrasing of the same goal)?}\\[5pt]

        \texttt{---}\\[3pt]
        \texttt{\#\# Output Requirement (STRICT)}\\
        \texttt{Return ONLY one JSON object (no extra text, no code fences):}\\
        \texttt{\{}\\
        \texttt{~~"score": <int>,~~// 0--100 total diversity score}\\
        \texttt{~~"subscores": \{}\\
        \texttt{~~~~"intent\_variety": <int>,~~// 0--25}\\
        \texttt{~~~~"action\_diversity": <int>,~~// 0--25}\\
        \texttt{~~~~"goal\_coverage": <int>,~~// 0--25}\\
        \texttt{~~~~"redundancy\_minimization": <int>~~// 0--25}\\
        \texttt{~~\},}\\
        \texttt{~~"analysis": "<short reasoning describing overall diversity and possible overlaps>",}\\
        \texttt{~~"representative\_examples": ["<one or two tasks that illustrate high or low diversity>"]}\\
        \texttt{\}}\\[5pt]

        \texttt{---}\\[3pt]
        \texttt{\#\# Additional Constraints}\\
        \texttt{- Evaluate ONLY diversity, not task quality or feasibility.}\\
        \texttt{- Consider whether tasks collectively span multiple distinct purposes, operations, or workflows.}\\
        \texttt{- Do NOT propose new tasks, rephrase them, or rewrite anything.}\\
        \texttt{- Do NOT remove or modify any input tasks.}\\
        \texttt{- The reasoning should briefly summarize what aspects contribute most to or detract from diversity.}\\[2pt]

        \vspace{1pt}
        \end{minipage}
    }
    \caption{Prompt for evaluating diversity of high-level tasks in a single web environment.}
    \label{tab:diversity-eval-prompt}
\end{table*}

\subsection{Baseline Implementation}
\label{app:explorer_analysis}
All baselines are re-implemented using GPT-4.1 for fair comparison. For Self-Instruct~\citep{wang2023selfinstructaligninglanguagemodels}, we follow the standard prompting approach to generate tasks from seed examples, as similar implementation by~\citet{sun2025osgenesisautomatingguiagent}. For OS-Genesis~\citep{sun2025osgenesisautomatingguiagent}, we adapt the generation part of the released codebase to WebArena, using random exploration to collect interaction triplets and generate tasks from single-step observations.

For Explorer~\citep{pahuja2025explorerscalingexplorationdrivenweb}, we believe our implementation is a faithful adaptation of the released codebase with minimal changes for WebArena compatibility (e.g., aligning browser operations). We additionally implement authentication handling, which the original Explorer does not support, to allow operation on login-gated pages. We have carefully investigated Explorer's poor performance: as shown in Table~\ref{tab:stat_complete_rate_traj}, the dominant failure mode arises from its overly aggressive refinement strategy ($8.6$ refinements per trajectory on average) that produces long trajectories and frequently exhausts the step budget ($68.3\%$ exceed the budget). Through trajectory inspection, we confirmed this behavior matches the algorithmic design rather than an implementation error. 

\subsection{Trajectory Quality Evaluation}
\label{app:judge_stats}
To ensure our quality assessments are not dependent on a single model, we evaluate synthesized data using three different judges: GPT-4.1, GPT-5.1, and Qwen3-30B~\citep{yang2025qwen3}. Results with GPT-5.1, GPT-4.1, and Qwen3-30B are shown in Table~\ref{tab:stat_data}, Table~\ref{tab:stat_data_gpt4}, and Table~\ref{tab:stat_data_qwen}, respectively.

Across all three different judges, \ours consistently produces higher-quality task-trajectory data compared to baselines. While absolute scores vary by judge, the relative ranking of methods remains stable, indicating that our improvements are robust across different evaluation models.

\subsection{Task Generation}
For \ours, the generation process begins by categorizing low-level actions into distinct functional groups through systematic exploration of the web environment. The agent interacts with the interface and analyzes all visible elements, distinguishing between interactive and non-interactive components and assigning each to an appropriate category. This structured categorization enables the model to generate contextually grounded low-level actions within different functional groups, thereby enhancing its generalization ability. The detailed prompt for this process is shown in Table~\ref{tab:cat-prompt}, which is largely adopted from OS-Genesis~\citep{sun2025osgenesisautomatingguiagent}. Using the low-level actions, we further generate high-level tasks following Table~\ref{tab:webarena-zero-shot-eval-prompt}.

\subsection{Trajectory Generation}
To ensure higher-quality trajectories, \ours divides the collection process into three stages.
In the first stage, given the set of collected high-level tasks, \ours executes them in a chain-of-thought (CoT) manner~\cite{wei2022chain}, as illustrated in Table~\ref{tab:cot-prompt}. The model reasons step-by-step over the current web context to generate the next low-level action until the task is completed or deemed infeasible.
During the second stage, as the agent interacts with the environment, each high-level task is dynamically analyzed and refined using the task refinement prompt (i.e., prompt at Table~\ref{tab:association-web-prompt}). \ours continuously evaluates whether the task remains valid given the current observation and refines it when inconsistencies, missing parameters, or execution stalls are detected.
In the final stage, \ours assesses the collected trajectories and refines them by dropping those deemed unachievable or reordering intermediate steps when a revised sequence better fulfills the task objective using prompt at Table~\ref{tab:y-refine-prompt}.

\subsection{Agent Framework}
Our agent execution framework builds upon ReAct~\citep{yao2023reactsynergizingreasoningacting}, instantiating a policy that interleaves CoT reasoning with environment actions in a browser-based setting with textual accessibility tree and visual screenshot as observations. Specifically, given a high-level instruction, the agent performs a sequence of steps to accomplish the task. At each step, the agent first takes observations from the environment based on the history actions and current state which includes the URL, accessibility tree, a set of candidate interactive elements and screenshot, then uses CoT reasoning to generate the next action to take. Execution continues until the model explicitly outputs a termination (``none'' action in Table~\ref{tab:action_table}), the task is judged unachievable by itself (``stop'' action in Table~\ref{tab:action_table}), or a preset step budget ($30$) is exhausted. The used prompt template is shown in Table~\ref{tab:cot-prompt}.

\subsection{Tasks Analysis}
We use LLMs-as-a-judge~\citep{llmasajudge} to evaluate the quality of the generated tasks. We prompt the LLM with Table~\ref{tab:diversity-eval-prompt} to assesses how broadly and non-redundantly the tasks cover different user intentions for specific website.

\section{Case Study}
\label{app:case_study}

\paragraph{Categorized Category Sets.}
To further illustrate the functional breadth captured by our categorized categorization, we list representative category sets for each website:
\begin{itemize}
  \item \textbf{Shopping}: Global Navigation, Search and Discovery, Shopping Options / Filtering, Sorting and Viewing Options, Product Browsing, Product Details and Purchase, Cart and Comparison, User Account Management, Product Reviews \& Feedback, Subscription.
  \item \textbf{CMS}: Global Navigation, User Account Management, Order Management, Product Management, Tab Navigation, Reporting \& Analytics, Search \& Filters, Table Controls \& Pagination, Admin Controls, Help and Information.
  \item \textbf{Reddit}: Navigation, User Account \& Profile, Notifications \& Submissions, Content Submission, Content Interaction (Posts \& Voting), Comments / Threading / Visibility, Moderation \& Toolbox, Forum Subscription \& Management, Search \& Discovery, Footer / Site Info.
  \item \textbf{Gitlab}: Menu \& Utility Actions, User/Account Actions, Repository High-Level Actions, Search \& Filtering, Commit History Navigation, Alerts and Notifications, Code Commenting/Review, Project Management, Navigation \& Site Structure, Issue Management.
  \item \textbf{Maps}: Authentication/Login, Account Registration, Map Editing, Navigation \& Page Switching, Sharing, Map Control, Community and Support, Map Data, Directions Input \& Management, Sidebar/Popup Controls.
\end{itemize}

Across these environments, our method consistently identifies meaningful, task-relevant functional groups (e.g., navigation, search, account management, content interaction), which supports the claim that categorized exploration is both expressive and robust enough to guide task synthesis beyond simple random exploration.

\paragraph{Task Refinement.}
Figure~\ref{fig:task-refine} illustrates an example of the task refinement process in \ours. The original task, ``Sort the 'vitamin supplements' search results by price to find the cheapest product available'', was initially executed based on the agent's understanding of the web interface. However, during interaction, the agent detected that the page failed to redirect to the intended subcategory ('Diet \& Sports Nutrition'), making the original goal inconsistent with the current observation. \ours automatically analyzed this discrepancy and refined the task to align with the accessible context, updating it to ``Identify the product with the lowest listed price in the 'Health \& Household' category.''. This refinement ensures that the high-level task remains executable and contextually valid for the existing trajectory.

\paragraph{Trajectory Refinement.}
Figure~\ref{fig:case-trajectory-refine} presents the trajectory refinement process in \ours. The agent's intermediate sequence of the trajectory for the task ``Find the cheapest available product in the Electronics category by sorting results by price'' contained significant noise and inefficiencies. During execution, the agent became stuck attempting to interact with a non-functional sort option, resulting in 19 repeated steps with no progress on the webpage. The collected trajectory included this futile loop along with redundant scroll actions that did not contribute to task completion. Our trajectory refinement step, equipped with the global view of the full trajectory and final task objective, identified and removed the repetitive clicking attempts on the unresponsive sort interface, and consolidated the necessary scroll actions. The refined trajectory retained only the essential 9 steps needed to successfully complete the task: clicking the Electronics category, navigating through sorting options, and scrolling to identify the cheapest product. This refinement ensures that the trajectory aligns precisely with the high-level task while removing noisy actions that would be harmful for the agent during fine-tuning.

\end{document}